\documentclass[conference]{IEEEtran}
\IEEEoverridecommandlockouts
\pdfoutput=1
\usepackage{cite}
\usepackage{amsmath,amssymb,amsfonts}
\usepackage{algorithmic}
\usepackage{graphicx}
\usepackage{textcomp}
\usepackage{xcolor}
\usepackage{booktabs}
\usepackage[T1]{fontenc}
\usepackage{subfigure}
\usepackage{graphicx}
\def\BibTeX{{\rm B\kern-.05em{\sc i\kern-.025em b}\kern-.08em
    T\kern-.1667em\lower.7ex\hbox{E}\kern-.125emX}}
\begin{document}

\title{GraphTEN: Graph Enhanced Texture Encoding Network}  
\author{  
\IEEEauthorblockN{Bo Peng$^1$, Jintao Chen$^2$, Mufeng Yao$^2$, Chenhao Zhang$^1$, Jianghui Zhang$^1$, Mingmin Chi$^{2*}$, Jiang Tao$^{1*}$}  
\IEEEauthorblockA{$^1$School of Information, Shanghai Ocean University, Shanghai\\
$^2$School of Computer Science, Fudan University, Shanghai\\
 \{bpeng,jtao\}@shou.edu.cn, \{m240751945, m240751960\}@st.shou.edu.cn, mfyao21@m.fudan.edu.cn, \{jtchen18,mmchi\}@fudan.edu.cn  
\thanks{$^*$Corresponding author.}}   
}

\maketitle

\begin{abstract}
Texture recognition is a fundamental problem in computer vision and pattern recognition. Recent progress leverages feature aggregation into discriminative descriptions based on convolutional neural networks (CNNs). However, modeling non-local context relations through visual primitives remains challenging due to the variability and randomness of texture primitives in spatial distributions. 
In this paper, we propose a graph-enhanced texture encoding network (GraphTEN) designed to capture both local and global features of texture primitives. GraphTEN models global associations through fully connected graphs and captures cross-scale dependencies of texture primitives via bipartite graphs. Additionally, we introduce a patch encoding module that utilizes a codebook to achieve an orderless representation of texture by encoding multi-scale patch features into a unified feature space. The proposed GraphTEN achieves superior performance compared to state-of-the-art methods across five publicly available datasets.

\end{abstract}


\section{Introduction}
Texture, as a fundamental visual attribute, encapsulates the spatial organization of basic elements within texture-rich images, serving as a vital representation of the underlying microstructure in natural scenes. Textured regions are typically characterized by repetitive patterns with inherent variability, making them essential pre-attentive visual cues for comprehending natural scenes. This unique property has enabled a wide range of applications, including medical image analysis, content-based image retrieval, and material classification~\cite{liu2019bow,peikari2015triaging}.
For decades, handcrafted texture descriptors have formed the basis of classical material and texture recognition methods. Techniques such as Gray-Level Co-occurrence Matrices (GLCM)\cite{4309314}, Local Binary Patterns (LBP)\cite{DBLP:journals/ejivp/KylbergS13}, and Gabor Filters~\cite{DBLP:journals/prl/IdrissaA02} have been widely utilized due to their simplicity and effectiveness. Further advancements introduced aggregation-based approaches such as Bag of Words (BoW) and Vector of Locally Aggregated Descriptors (VLAD), which enhanced the representation of texture descriptors by leveraging visual codebooks and residual vector aggregation~\cite{DBLP:conf/cvpr/JegouDSP10}.
\begin{figure}[ht]
  \centering
  \includegraphics[width=\linewidth]{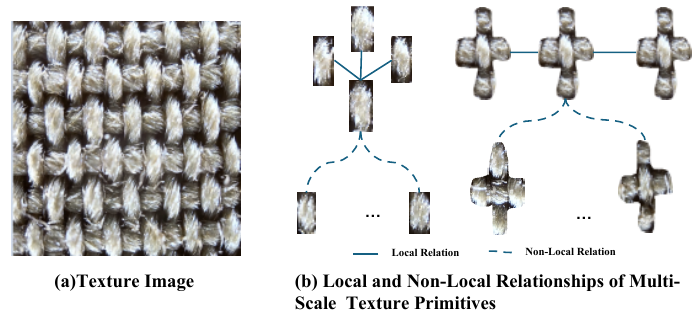}
  \caption{Texture primitives demonstrate significant local and global associations across multiple scales, with solid lines representing local dependencies and dashed lines capturing non-local relationships. }
  \label{idea}
    \vspace{-2em}

\end{figure}

With the rise of deep learning, convolutional neural networks (CNNs) have emerged as the dominant framework for texture recognition. Methods such as FV-CNN~\cite{liu2019bow}, Deep Texture Encoding Network (DeepTEN)\cite{DBLP:conf/cvpr/ZhangXD17}, and DSRNet\cite{zhai2020deep} leverage CNN representations through orderless aggregation or spatial dependency modeling to extract texture features effectively. More recently, fractal-based approaches, including CLASSNet~\cite{chen2021deep} and FENet~\cite{xu2021encoding}, have incorporated multi-scale fractal analysis to represent textures, enabling better adaptability to spatial distributions and variability.

Building upon these foundations, recent research has continued to advance texture representation and recognition by exploring innovative perspectives. For example, Zhu et al.\cite{zhu2023learning} propose a learnable Gabor-based framework that integrates trainable statistical feature extractors with deep neural networks to enhance fine-grained texture recognition. Similarly, Mohan et al.\cite{mohan2024lacunarity} introduce a lacunarity pooling layer that effectively captures spatial heterogeneity and distributions, demonstrating notable improvements in plant texture classification. Furthermore, Zhai et al.~\cite{zhai2023exploring} develop a unified framework that simultaneously models structural relationships and attribute associations in natural textures, contributing to a more comprehensive understanding of texture analysis.
While these methods excel in handling repetitive and periodic texture patterns, real-world applications often defy these assumptions. Texture primitives frequently exhibit non-uniform arrangements, distortions, and deformations, challenging models reliant on regularity.

 \begin{figure*}[ht]
\centering
\includegraphics[width=\textwidth]{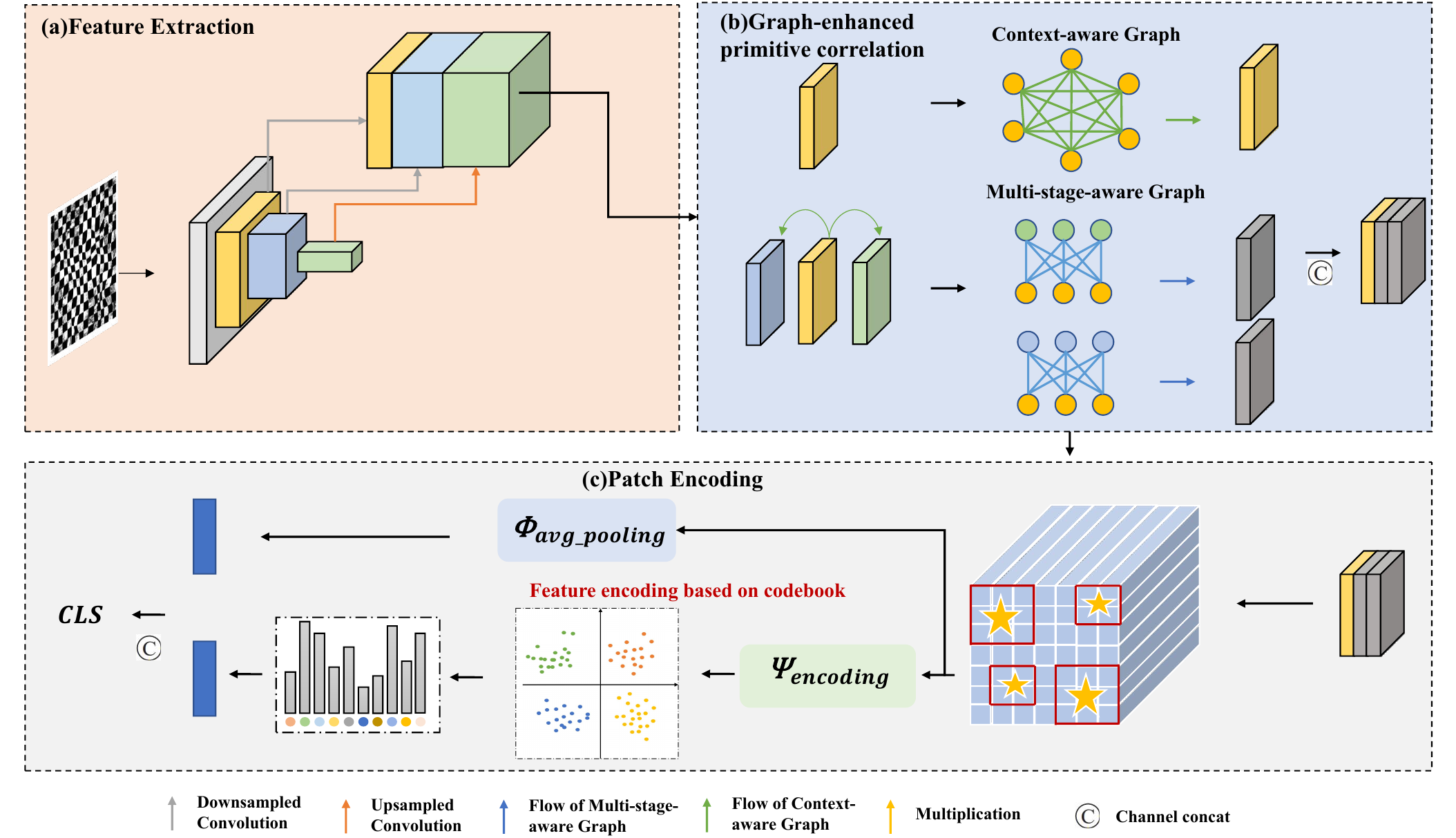}
\caption{Overall architecture of the proposed Graph enhanced texture encoding network (GraphTEN).}
\label{model}
  \vspace{-1em}

\end{figure*}

To address the aforementioned limitations, we propose GraphTEN, a graph neural network-based texture encoding framework tailored for robust texture representation. At its core, GraphTEN transforms CNN feature maps into a graph structure and leverages dual-directional graph networks to enhance feature extraction by incorporating both contextual and multi-stage awareness. Additionally, we design a Patch Encoding Module (PE) to sample multi-scale features and encode them into a unified feature space, enabling seamless local-global feature aggregation for comprehensive texture analysis.

The main contributions of this paper are summarized as follows:

\begin{itemize}  
\item We propose a novel graph-enhanced texture encoding network (GraphTEN) that captures both local and global texture features. By modeling global associations using fully connected graphs and cross-scale dependencies via bipartite graphs, GraphTEN effectively addresses the variability of texture primitives.  
\item A Patch Encoding Module (PE) is introduced to encode multi-scale patch features into a unified feature space using a codebook, achieving an orderless representation and seamless local-global feature aggregation.  
\end{itemize}

\section{Related Work}  
\subsection{Texture Recognition}  
Texture recognition is applied across fields such as biomedical imaging, material recognition, and anomaly detection. Robust image descriptions are essential, capturing texture patterns across scales, positions, and rotations. Traditional methods rely on pooling local features like filter bank textons~\cite{leung2001representing,varma2005statistical} and image patches~\cite{varma2008statistical,varma2003texture}. Specialized descriptors~\cite{2016Facilitating,2016Sparse} target dynamic textures~\cite{2016Equiangular} and fine-grained materials. Gatys et al.~\cite{2015Texture} demonstrated the use of Gram matrices from VGG layers for texture synthesis.  
Lin et al.~\cite{lin2015bilinear} introduced bilinear CNNs, outperforming Fisher Vectors with efficient gradient computation for fine-tuning. Texture CNN (T-CNN)~\cite{andrearczyk2016using} utilized learnable filter banks with energy layers for feature pooling, while Wavelet CNN~\cite{fujieda2017wavelet} formulated convolution and pooling as frequency-domain filtering. LSCTN~\cite{bu2019deep} incorporated a locality-aware coding layer for joint dictionary and representation learning. Recently, Zhu et al.~\cite{zhu2023learning} proposed a learnable Gabor-based framework for fine-grained texture recognition, and Mohan et al.~\cite{mohan2024lacunarity} introduced lacunarity pooling for capturing spatial heterogeneity, yielding significant improvements. Zhai et al.~\cite{zhai2023exploring} developed a unified framework modeling structural relationships and attribute associations, advancing texture analysis comprehensively.

\section{Methodology}
\subsection{Graph-enhanced Primitive Correlation Module}

\textbf{Context-aware Graph.} 
Firstly, we perform graph computation within the same feature map stage using a fully-connected Graph Neural Network (GNN). Let $x = \left \{ x_{i} \in \mathbb{R}^{C}, i=1,...,N \right \}$ be a set of convolutional features, where $x_i \in C$ represents the feature vectors from a feature map stage, and $i$ indexes the spatial location of each feature vector. For example, given an image feature map defined on a 2D grid with dimensions $h \times w \times c$, where $h$, $w$, and $c$ denote the height, width, and number of channels, respectively, we can represent $N = h \times w$. 
We then construct a fully-connected graph $\mathcal{G} = (\mathcal{V}, \mathcal{E})$ with $N$ nodes $v_i \in \mathcal{V}$ and edges $(e_i, e_j) \in \mathcal{E}$ to represent the relationships between nodes. The GNN assigns $x_i$ as the input to the node $v_i$ and computes the feature representation of each node through an iterative message-passing process. The propagation process of the graph layer can be expressed as:

\begin{equation}
    \mathbf{\hat{G}}= \mathbf{F_{GPC}}(\mathbf{G}) =  \sigma \left( \frac{1}{\mu(\mathbf{X})} \sum_{j=1}^{N} g(\mathbf{v_i},\mathbf{v_j}) \right),
\end{equation}

where $\hat{G}$ represents the updated feature representation at node $i$, $\sigma$ is an element-wise activation function, $g(\cdot)$ is a kernel function encoding the relations between two feature vectors $G_i$ and $G_j$ (as suggested by \cite{wang2018non}, $g$ is chosen as a Gaussian function), and $\mu$ is a normalization factor for node $i$. 

\textbf{Multi-stage-aware Graph.} 
To effectively model the relationships across different feature map stages, we construct a bipartite graph using the convolutional feature maps. Let $\mathcal{F}_{0}^{l}$ and $\mathcal{F}_{1}^{l}$ represent the feature maps from two different stages at layer $l$, each with a resolution of $H_l \times W_l$. Each feature map $\mathcal{F}_{m}^{l}$, where $m \in \{0, 1\}$, contains $C$ channels, representing the output from the convolutional layers.

In this context, each pixel $(u, v)$ within $\mathcal{F}_{0}^{l}$ and $\mathcal{F}_{1}^{l}$ is treated as a node within the graph. The feature vector at location $(u, v)$ in $\mathcal{F}_{m}^{l}$ is denoted as $\mathbf{x}_{m}[u, v] \in \mathbb{R}^{C}$, where $C$ is the number of channels. We then construct a fully-connected bipartite graph $\mathcal{G} = (\mathcal{V}, \mathcal{E})$ with nodes $v_{i} \in \mathcal{V}$ and edges $(e_{i,j}) \in \mathcal{E}$, where each edge connects nodes between the two stages $\mathcal{F}_{0}^{l}$ and $\mathcal{F}_{1}^{l}$.

The adjacency matrix $\mathcal{G}$ of this bipartite graph is four-dimensional, denoted as $\mathcal{G} \in \{0, 1\}^{H_l \times W_l \times H_l \times W_l}$. The element $\mathcal{G}[u, v, i, j]$ is set to 1 if there is an edge between nodes $(u, v)$ from $\mathcal{F}_{0}^{l}$ and $(i, j)$ from $\mathcal{F}_{1}^{l}$, and 0 otherwise.

To account for the variability and potential semantic differences across stages, we encode the edges of the graph based on the contextual similarity between pixels. This is done by first generating context features $\mathcal{F}_{0,c}^{l}$ and $\mathcal{F}_{1,c}^{l}$ using non-parameterized dilated convolutions with dilation factor $d$ and kernel size $k$. The context feature at each pixel $(u,v)$ is computed as follows:

\begin{equation}
\mathcal{F}_{m,c}^{l}[u,v] = \frac{1}{k^2 - 1}\sum_{(i,j)\in [-k, k]^2} \mathcal{F}_{m}^{l}[u + i \cdot d, v + j \cdot d]
\end{equation}

where $m \in \{0,1\}$, and the summation is performed over a $k \times k$ neighborhood around pixel $(u, v)$.

For each pixel $(u,v)$ in $\mathcal{F}_{0,c}^{l}$, the top-$n$ nearest neighbors in $\mathcal{F}_{1,c}^{l}$ are identified based on their Euclidean distance:

\begin{equation}
\mathcal{H}_{u,v} = \textbf{TOPK}_{(i,j) \in [0, H_l] \times [0, W_l]}\left( \left \| \mathcal{F}_{0,c}^{l}[u,v] - \mathcal{F}_{1,c}^{l}[i, j] \right \|_2 \right)
\end{equation}

The adjacency matrix $\mathcal{G}$ is then defined as:

\begin{equation}
\mathcal{G}[u,v,i,j] = \begin{cases}
1, & \text{if } (i,j) \in \mathcal{H}_{u,v} \\
0, & \text{otherwise}
\end{cases}
\end{equation}

The resulting graph feature representation for this multi-stage-aware graph is denoted as $\hat{G'}$. 

Finally, given the features output by the context-aware graph layer $\hat{G}$ and the multi-stage-aware graph layer $\hat{G'}$, we build texture-aware structure features by applying a fusion function formulated as:

\begin{equation}
 V = \mu \left( g\left( \text{Concat} \left[ \hat{G}; \hat{G'} \right] \right) \right),
\end{equation}

where $g(\cdot)$ is a convolution layer and $\mu$ denotes a sigmoid activation.

\subsection{Patch Encoding Module}


As illustrated in Fig.~\ref{model}, hierarchical features are first extracted using a CNN-based backbone, resulting in a 3D feature map \( Q \in \mathbb{R}^{W \times H \times C} \). To capture multi-scale local information, multi-scale sliding windows are applied to densely sample sub-regions from \( Q \). A set of window sizes \( W = \{d_1, d_2, \dots, d_{N_w}\} \) is defined, where \( N_w \) is the number of patch sizes and \( d \) represents each size. For each \( d \) in \( W \), patches \( P = \{P_i \in \mathbb{R}^{d \times d \times C}, i = 1, \dots, N_p\} \) are extracted from \( Q \), with the total number of patches given by:  

\begin{equation}  
N_{p}=\left \lfloor \frac{H-d}{s}+1 \right \rfloor\ast \left \lfloor \frac{W-d}{s}+1 \right \rfloor ,d\leq H,d\leq W,  
\end{equation}  

where \( s \) is the sliding window stride. For each patch \( p_i \in \mathbb{R}^{d \times d \times C} \) in \( P \), a texture encoding function \( \mathit{\Psi}_{encoding} \), inspired by~\cite{DBLP:conf/cvpr/ZhangXD17,DBLP:conf/cvpr/Xue0D18}, computes a soft histogram \( H \in \mathbb{R}^{K \times D} \), expressed as:  

\begin{equation}  
 H = \mathit{\Psi}_{encoding}(p_i),  
\end{equation}  

where \( p \) is represented by \( V = \{ V_{n_{v}} \in \mathbb{R}^{d \times d}, n_{v}=1, \dots, C \} \), and a learnable codebook \( c = \{ c_k \in \mathbb{R}^{D}, k=1, \dots, K \} \) defines cluster centers. Each \( H_k \in H \) is computed by applying softmax to the residual error vector:  

\begin{equation}  
    H_k =\sum_{i=1}^{N}\frac{\exp(-s_{k}||V_i-c_k||^2)}{\sum_{j=1}^{K}\exp(-s_{j}||V_j-c_j||^2)},  
\end{equation}  

where \( s \) is a smoothing factor for each cluster center \( c_k \), and \( H_k \in \mathbb{R}^{D} \).  

The multi-scale patch feature maps are then encoded using \( \mathit{\Psi}_{encoding} \), with different weights assigned to features from different scales. The overall encoding process is expressed as:  

\begin{equation}  
U = \sum_{j=1}^{N_{w}}\sum_{i=1}^{N_{p}}w_j\mathit{\Psi}_{encoding}(P_{i}),  
\end{equation}  

where \( N_{w} \) is the number of scales in \( W = \{d_{j}\}_{j=1}^{N_{w}} \), \( N_p \) is the number of patches in \( P = \{P_{i}\}_{i=1}^{N_{p}} \) for each \( d_j \), and \( w_j \) is the weight for scale \( d_j \). Finally, global average pooling is applied to the original feature map \( Q \in \mathbb{R}^{W \times H \times C} \) to obtain a low-dimensional feature, which is concatenated with \( U \) to form the fusion representation \( Z \) for texture recognition:  

\begin{equation}  
    Z = Concat\left[ \mathit{\Phi}_{avg\_pooling}(Q); \mathit{\Phi}_{flatten}(U) \right].  
\end{equation}

\begin{table*}[htbp]
  \centering
  \caption{Performance comparison of different methods in terms of classification accuracy (\%).}
    \begin{tabular}{lrrrrrrrrrrrr}
    \hline
          & \multicolumn{2}{c}{DTD} & \multicolumn{2}{c}{MINC} & \multicolumn{2}{c}{FMD} & \multicolumn{2}{c}{Fabrics} & \multicolumn{2}{c}{GTOS} & \multicolumn{2}{c}{KTH} \\
    Method & \multicolumn{1}{c}{mean} & \multicolumn{1}{c}{std} & \multicolumn{1}{c}{mean} & \multicolumn{1}{c}{std} & \multicolumn{1}{c}{mean} & \multicolumn{1}{c}{std} & \multicolumn{1}{c}{mean} & \multicolumn{1}{c}{std} & \multicolumn{1}{c}{mean} & \multicolumn{1}{c}{std} & \multicolumn{1}{c}{mean} & \multicolumn{1}{c}{std} \\
    \hline
    FC-CNN(CVPR15)\cite{cimpoi2015deep} & 62.9  & 0.8   & 60.4  & 0.5   & 77.5  & 1.8   & 57.9  & 0.6   & 68.5  & 0.6   & 81.8  & 2.5 \\
    FV-CNN(CVPR15)\cite{cimpoi2015deep} & 72.3  & 1     & 69.8  & 0.5   & 79.8  & 1.8   & 66.5  & 0.9   & 77.1  & 0.6   & 75.4  & 1.5 \\
    BCNN(CVPR16)\cite{lin2016visualizing}  & 69.6  & 0.7   & 67.1  & 1.1   & 77.8  & 1.9   & 65.6  &  -     & 78.7  & 0.3   & 75.1  & 2.8 \\
    Deep-TEN(CVPR17)\cite{DBLP:conf/cvpr/ZhangXD17} & 69.6  & 0.5   & 81.3  & 0.7   & 80.2  & 0.9   & 75.2  & 0.7   & 84.5  & 0.4   & 82    & 3.3 \\
    DEP(CVPR18)\cite{DBLP:conf/cvpr/Xue0D18}   & 73.2  & 0.5   & 82    & 0.7   & 80.7  & 0.7   & 74.3  & 1.2   &-       &-       & 82.4  & 3.5 \\
    MAPNet(ICCV19)\cite{zhai2023exploring} & 76.1  & 0.6   &-       & -      & 85.2  & 0.7   &-       &-       & 84.7  & 2.2   & 84.5  & 1.3 \\
    DSRNet(CVPR20)\cite{zhai2020deep} & 77.6  & 0.6   & -      &-       & 86    & 0.8   &-       &-       & 85.3  & 2     & 85.9  & 1.3 \\
    HistNet(PR21)\cite{peeples2020histogram} & 72    & 1.2   & 82.4  & 0.3   &-       & -      & -      &-       &-       &-       &-       &-  \\
    FENet(NeurIPS21)\cite{xu2021encoding} & 74.2    & 0.1   & 83.9      &0.1       & 86.7      & 0.2       &-       & -      & 85.7      & 0.1      & 88.2      & 0.2 \\
    CLASSNet(CVPR21)\cite{chen2021deep} & 74    & 0.5   & 84    & 0.6   & 86.2  & 0.9   & -      &-       & 85.6  & 2.2   & 87.7  & 1.3 \\
    MPAP(TPAMI24)\cite{zhai2023exploring} & 78    & 0.5   & 82.5    & 0.1   & 87.6  & 0.9   &-       &-       & 86.1  & 1.8   & 87.9  & 1.5 \\
    GraphTEN(ours) & \textbf{79.1}  & 0.6   & \textbf{85.2}  & 0.3   & \textbf{87.7}  & 1.2   & \textbf{80.7}  & 0.6   & \textbf{86.8}  & 2.5   & \textbf{87.4}  & 1.6 \\
    \hline
    \end{tabular}%
  \label{result1}%
\end{table*}%
\section{Experiments}

\subsection{Experimental Setting}  
\textbf{Datasets.} The proposed method is evaluated on six widely-used benchmark datasets. The Describable Textures Database (DTD)~\cite{cimpoi14describing} comprises 47 texture categories, each containing 120 images, with ten predefined splits for training, validation, and testing. The Flickr Material Dataset (FMD)~\cite{sharan2013recognizing} consists of ten material categories and is a standard benchmark for material classification. The Materials in Context Database (MINC)~\cite{bell2015material} includes 23 material classes, with 2500 images per class, and provides five training/testing splits. The Fabrics dataset~\cite{Eccv16KampourisZGM} serves as a publicly available resource for fine-grained material classification. Ground Terrain in Outdoor Scenes (GTOS)~\cite{xue2017differential} consists of 40 outdoor ground material classes, with a predefined training/testing split. Finally, the KTH-TIPS2b~\cite{caputo2005class} dataset includes texture-rich images from 11 material categories, captured under various conditions to simulate realistic scenarios.  

\textbf{Implementation.} To ensure fairness in comparisons, all experiments utilize ResNet-50 as the backbone network. GraphTEN is trained using cross-entropy loss and optimized with momentum stochastic gradient descent (SGD). The training process is conducted for 50 epochs with a mini-batch size of 64. Following empirical settings in prior work on fractal analysis~\cite{DBLP:conf/cvpr/Xue0D18,chen2021deep,DBLP:conf/cvpr/ZhangXD17}, the parameter \( K = 16 \) is used in the patch encoding module. The sliding window stride is set to 1, and square patch sizes of \{3, 5, 7\} are employed with associated weights of \( w_i = \{0.35, 0.45, 0.2\} \), respectively.  
For fine-tuning, the learning rate is initialized at 0.004 and reduced by a factor of 10 when the error plateaus. A weight decay of 0.001 and a momentum of 0.9 are applied to regularize and stabilize training.

\begin{table}[h]  
  \centering  
  \caption{Complexity comparison of different models in terms of  
number of model parameters (Params) and floating-point operations per second (FLOPs).}  
\resizebox{0.48\textwidth}{!}{  
    \begin{tabular}{cccccc}  
    \hline  
      Param  & ResNet50 & DeepTEN & DEPNet & CLASSNet & \textbf{Ours} \\
    \hline  
    Param($\approx$,M) & 23.57 & 23.91 & 25.56 & 23.7  & 26.12  \\
    FLOPS($\approx$,G) & 4.11  & 4.12  & 4.11  & 4.14  & 4.17 \\
    \hline  
    \end{tabular}}  
  \label{paras}%
    \vspace{-2em}  
\end{table}

\subsection{Experimental Results}

\textbf{Comparison with State-of-the-Art Methods.}   
As shown in Tab.~\ref{result1}, GraphTEN outperforms all benchmarks in recognition accuracy, achieving state-of-the-art results across five material/texture recognition datasets: DTD, MINC-2500, FMD, Fabrics, and GTOS, without relying on data augmentation.  

Specifically, on two large-scale datasets, DTD and MINC, GraphTEN surpasses prior methods by significant margins, demonstrating improvements of 1.5\% and 1.2\%, respectively. These results highlight the strength of combining regional and global feature representations, as GraphTEN effectively leverages a graph-based approach to connect local image primitives at a global scale.   

On the KTH-TIPS2b dataset, GraphTEN performs slightly below CLASSNet and FENet. However, for the small-scale dataset FMD, despite being constrained by limited training data, our method still achieves state-of-the-art results.   

\textbf{Fine-Grained Material Classification.}   
The Fabrics dataset~\cite{Eccv16KampourisZGM} is a challenging benchmark for fine-grained material classification, composed of over 2000 images of fabric surfaces, including visually similar materials (e.g., cotton, linen, and polyester). From the results in Table~\ref{result1}, it is evident that recent methods such as DeepTEN and DEP struggle on this dataset, while our model achieves a 5\% improvement over the previous state-of-the-art. This demonstrates the efficacy of our graph-based framework in capturing subtle distinctions in fine-grained material classes by effectively connecting local primitives in texture images.   

\textbf{Model Complexity and Efficiency.}   
We compare the complexity of GraphTEN against backbone CNNs, DeepTEN, DEPNet, and CLASSNet in terms of the number of parameters and floating point operations per second (FLOPs). As shown in Tab.~\ref{paras}, GraphTEN achieves competitive complexity metrics compared to other models across both criteria, with no significant increase in parameter count or computational cost.

\begin{table}[h!]
  \centering
  \caption{Ablation experiment for the proposed modules.}
  \resizebox{0.5\textwidth}{!}{
    \begin{tabular}{lcccc}
    \hline
    Method & DTD   & MINC  & FMD   & Fabrics \\
    \hline
    FE                        & 65.4  & 70.2  & 77.9  & 63.2 \\
    ViG\cite{han2022vision}                       & 67.6  & 71.4  & 73.1  & 65.5 \\
    Vit\cite{dosovitskiy2020image}                       & 69.2  & 72.5  & 79.2  & 69.7 \\
    Swin-T\cite{liu2021swin}                    & 65.4  & 73.4  & 81.9  & 70.3 \\
    FE+ViG\cite{han2022vision}                    & 66.9  & 70.8  & 78.4  & 67.1 \\
    FE+Swin\cite{liu2021swin}                   & 71.7  & 73.9  & 77.9  & 71.6 \\
    FE+Non-local\cite{wang2018non}              & 73.5  & 76.7  & 82.2  & 72.7 \\
    FE+CAG                    & 74.5  & 82.7  & 84.8  & 75.8 \\
    FE+CAG+MAG                & 77.8  & 84.5  & 86.1  & 77.9 \\
    FE+CAG+MAG+PE         & 79.1  & 85.2  & 87.7  & 80.7 \\
    \hline
    \end{tabular}}%
  \label{abl}%
\end{table}%
\textbf{Ablation Study.}  
In this section, we evaluate the effectiveness of each proposed module through a series of ablation experiments. Tab.~\ref{abl} summarizes the results on our dataset under various settings. The components analyzed include the Feature Extractor (FE), Context-Aware Graph Layer (CAG), Multi-Stage-Aware Graph Layer (MAG), and Patch Encoding Module (PE). Additionally, for comparative purposes, we incorporate other global correlation modeling methods, such as ViT~\cite{dosovitskiy2020image}, ViG~\cite{han2022vision}, Swin~\cite{liu2021swin}, and Non-local~\cite{wang2018non}, into the ablation experiments.   

The results demonstrate that the Graph-Enhanced Primitive Correlation Module (comprising CAG and MAG) is pivotal in all tasks, significantly improving performance by effectively associating similar texture primitives and enhancing non-local representation capabilities. Similarly, the Patch Encoding (PE) layer contributes substantially by aggregating local features into a unified representation, thereby capturing both local and global texture characteristics.   
Importantly, when compared to other GNN or Transformer-based modules, our approach consistently achieves superior performance in texture representation, highlighting the advantages of leveraging graph-enhanced correlation modeling for texture recognition tasks.

\begin{figure}[h!]
  \centering
    \subfigure[DeepTEN] 
    {
    	\begin{minipage}[t]{0.24\textwidth}
    	\centering          
    	\includegraphics[width=\textwidth]{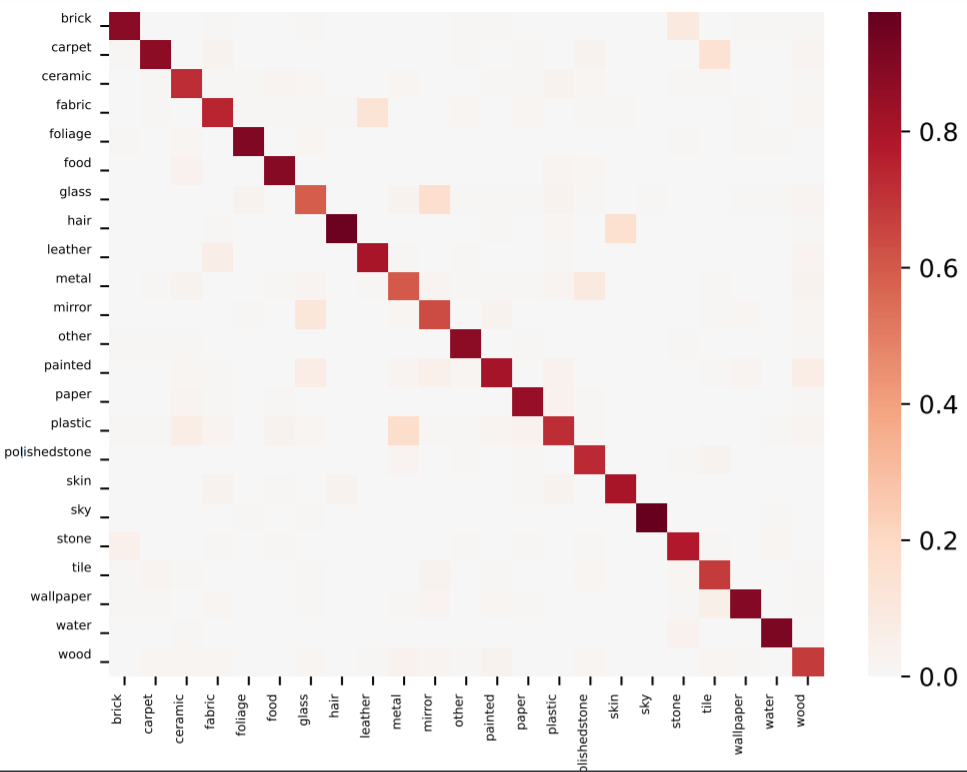}   
    	\label{fig:swmsa}
    	\end{minipage}
    }\subfigure[FENet] {
    	\begin{minipage}[t]{0.24\textwidth}
    	\centering          
    	\includegraphics[width=\textwidth]{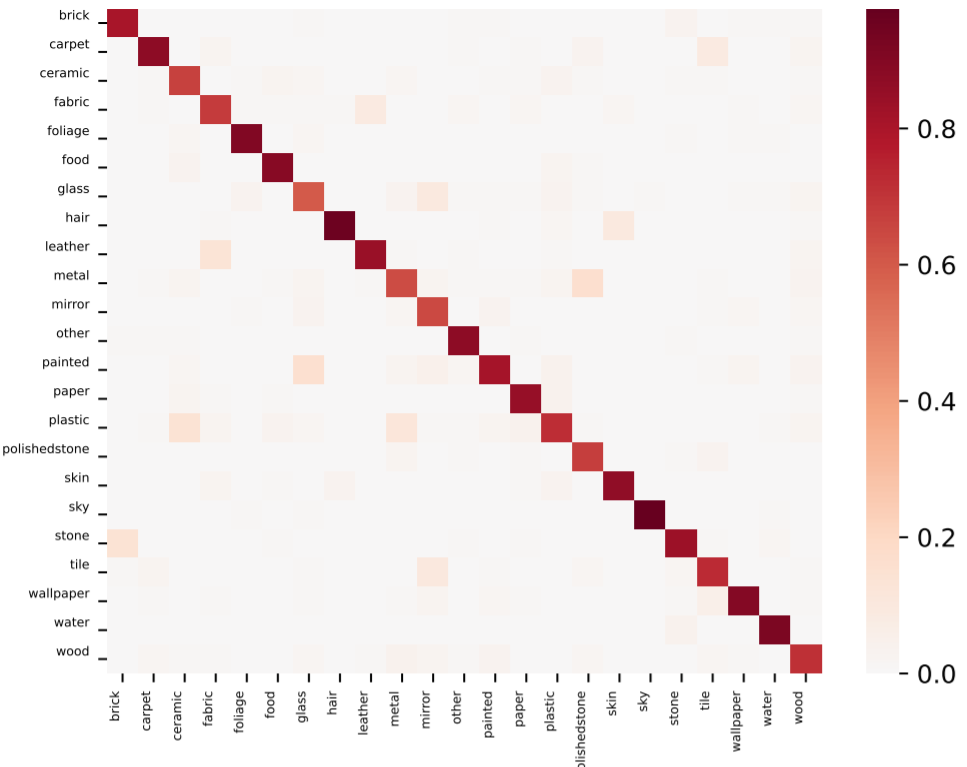}   
    	\label{fig:msa}
    	\end{minipage}
    }
    \subfigure[MPAP]{
    	\begin{minipage}[t]{0.24\textwidth}
    	\centering          
    	\includegraphics[width=\textwidth]{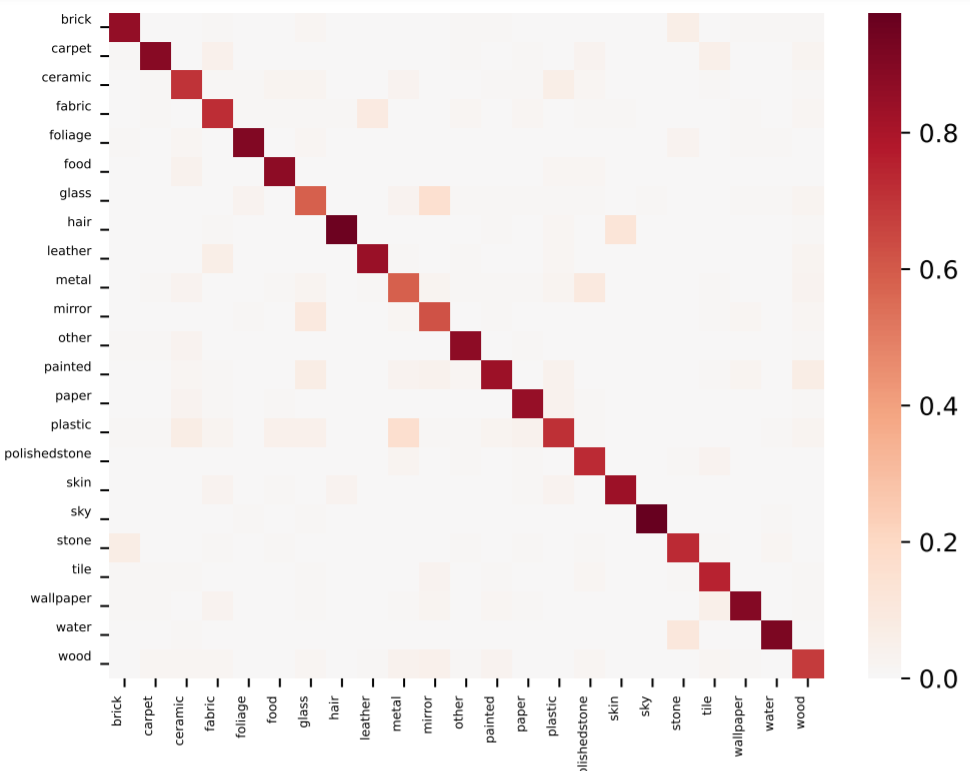}   
    	\label{fig:combine}
    	\end{minipage}
    }\subfigure[Ours] 
    {
    	\begin{minipage}[t]{0.24\textwidth}
    	\centering          
    	\includegraphics[width=\textwidth]{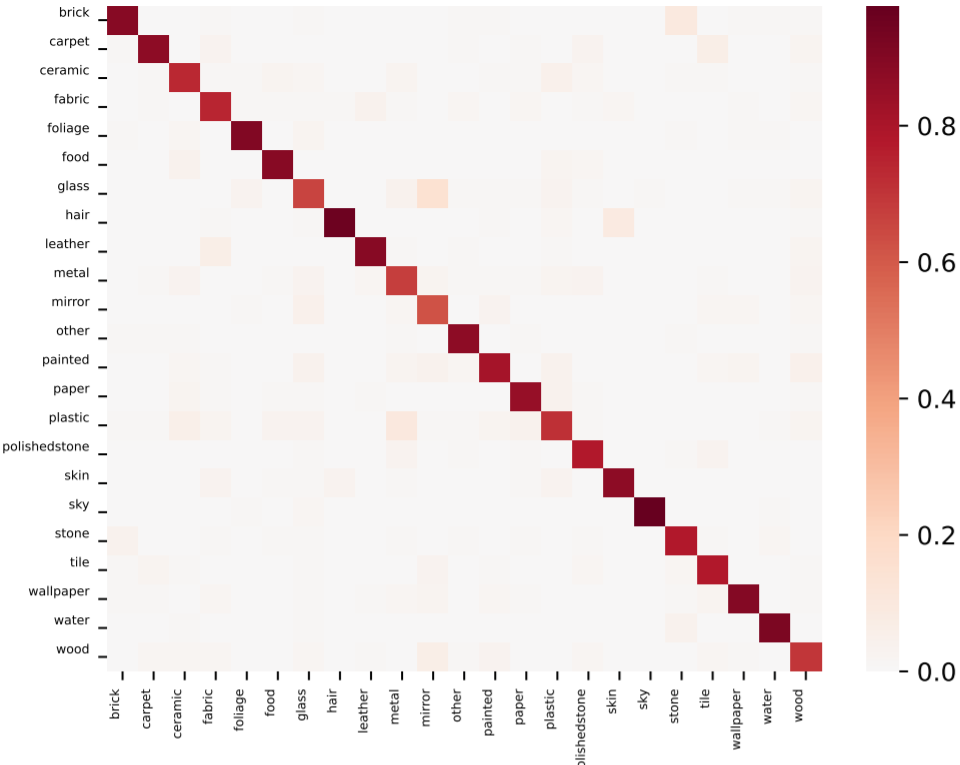}   
    	\label{fig:combine}
    	\end{minipage}
    }
  \caption{ Comparing our method with benchmarks via confusion matrix.}
  \label{confusion}
    \vspace{-1em}

\end{figure}

\begin{figure}[h!]
  \centering
    \subfigure[ Samples confusing of ResNet] 
    {
    	\begin{minipage}[t]{0.48\textwidth}
    	\centering          
    	\includegraphics[width=\textwidth]{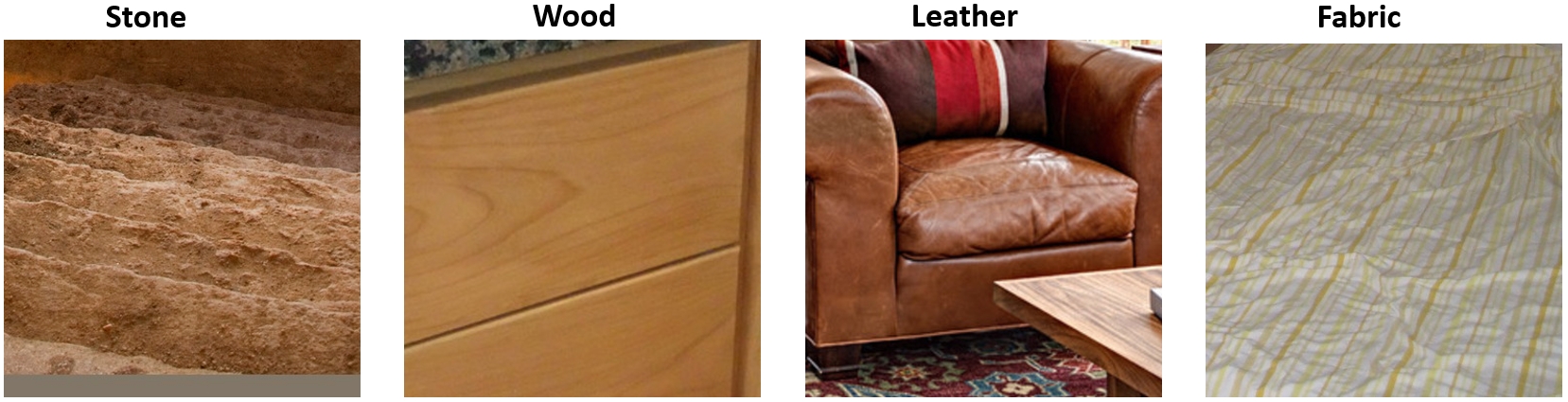}   
    	\label{fig:swmsa}
    	\end{minipage}
    }
    \subfigure[Samples confusing of GraphTEN(ours)] {
    	\begin{minipage}[t]{0.48\textwidth}
    	\centering          
    	\includegraphics[width=\textwidth]{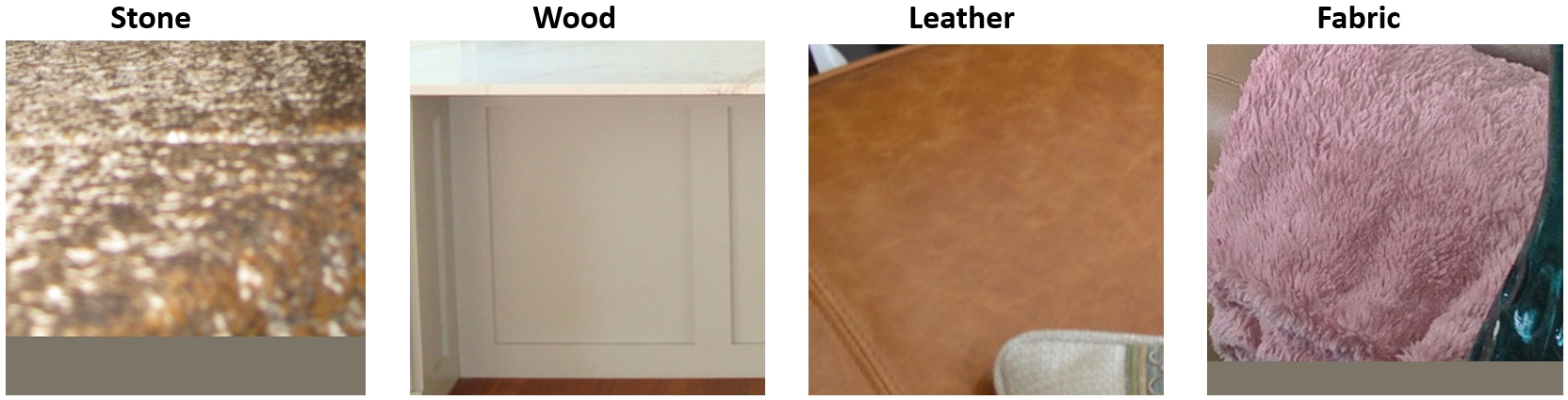}   
    	\label{fig:msa}
    	\end{minipage}
    }
    \subfigure[Example image and correct category] {
    	\begin{minipage}[t]{0.48\textwidth}
    	\centering          
    	\includegraphics[width=\textwidth]{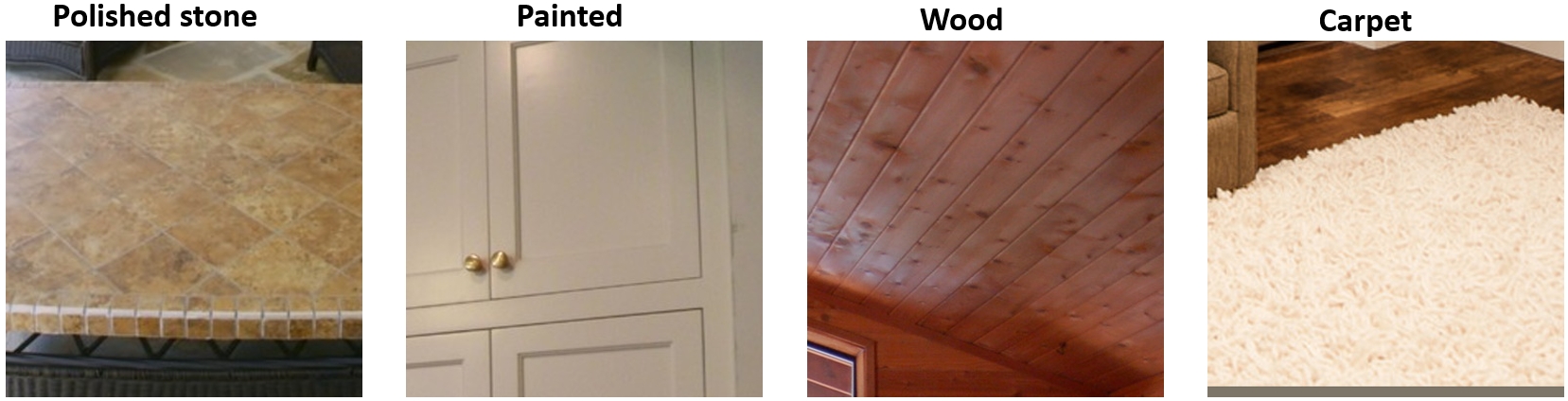}   
    	\label{fig:msa}
    	\end{minipage}
    }
  \caption{  Confusion analysis. The samples at the top row are incorrectly classified into the class of the corresponding samples at the bottom rows.}
  \label{confusion2}
    \vspace{-1.5em}

\end{figure}
\textbf{Study on Confusion Cases.}  
The confusion matrices on the test set of MINC-2500 (Fig.~\ref{confusion}) reveal that GraphTEN outperforms FENet, DeepTEN, and MPAP, offering fewer misclassifications and higher overall accuracy. As shown in Fig.~\ref{confusion2}(a), GraphTEN successfully classifies samples that are misclassified by the ResNet backbone. For example, while "Stone" and "Polished Stone" exhibit highly similar texture primitives, GraphTEN accurately distinguishes between the two by capturing fine-grained texture details via global association modeling. Similarly, GraphTEN effectively differentiates between closely related categories such as "Wood," "Leather," "Fabric," and "Carpet."  

However, as illustrated in Fig.~\ref{confusion2}(b), GraphTEN still makes errors in specific cases, such as misclassifying "Polished Stone" as "Stone," "Wood" as "Painted," "Leather" as "Wood," and "Fabric" as "Carpet." These misclassifications typically occur when the textures of the samples are nearly indistinguishable, even to humans. Despite these challenges, the confusion matrix analysis indicates that GraphTEN demonstrates superior texture and material recognition capabilities compared to existing methods.  

\section{Conclusion}  
This work presents the Graph-Enhanced Texture Encoding Network (GraphTEN), a novel framework that captures non-local correlations to advance texture recognition. By integrating graph-based modeling, GraphTEN effectively combines local texture primitives with global structural information, achieving state-of-the-art performance on five benchmark datasets.
Future work will investigate integrating multimodal models for broader texture understanding and developing lightweight graph architectures to enhance computational efficiency, paving the way for scalable real-world applications.

\vfill\pagebreak


\begin{thebibliography}{10}
\providecommand{\url}[1]{#1}
\csname url@samestyle\endcsname
\providecommand{\newblock}{\relax}
\providecommand{\bibinfo}[2]{#2}
\providecommand{\BIBentrySTDinterwordspacing}{\spaceskip=0pt\relax}
\providecommand{\BIBentryALTinterwordstretchfactor}{4}
\providecommand{\BIBentryALTinterwordspacing}{\spaceskip=\fontdimen2\font plus
\BIBentryALTinterwordstretchfactor\fontdimen3\font minus \fontdimen4\font\relax}
\providecommand{\BIBforeignlanguage}[2]{{%
\expandafter\ifx\csname l@#1\endcsname\relax
\typeout{** WARNING: IEEEtran.bst: No hyphenation pattern has been}%
\typeout{** loaded for the language `#1'. Using the pattern for}%
\typeout{** the default language instead.}%
\else
\language=\csname l@#1\endcsname
\fi
#2}}
\providecommand{\BIBdecl}{\relax}
\BIBdecl

\bibitem{DBLP:conf/cvpr/CimpoiMV15}
\BIBentryALTinterwordspacing
M.~Cimpoi, S.~Maji, and A.~Vedaldi, ``Deep filter banks for texture recognition and segmentation,'' in \emph{{IEEE} Conference on Computer Vision and Pattern Recognition, {CVPR} 2015, Boston, MA, USA, June 7-12, 2015}.\hskip 1em plus 0.5em minus 0.4em\relax {IEEE} Computer Society, 2015, pp. 3828--3836. [Online]. Available: \url{https://doi.org/10.1109/CVPR.2015.7299007}
\BIBentrySTDinterwordspacing

\bibitem{peikari2015triaging}
M.~Peikari, M.~J. Gangeh, J.~Zubovits, G.~Clarke, and A.~L. Martel, ``Triaging diagnostically relevant regions from pathology whole slides of breast cancer: A texture based approach,'' \emph{IEEE transactions on medical imaging}, vol.~35, no.~1, pp. 307--315, 2015.

\bibitem{4309314}
R.~M. {Haralick}, K.~{Shanmugam}, and I.~{Dinstein}, ``Textural features for image classification,'' \emph{IEEE Transactions on Systems, Man, and Cybernetics}, vol. SMC-3, no.~6, pp. 610--621, 1973.

\bibitem{DBLP:journals/ejivp/KylbergS13}
\BIBentryALTinterwordspacing
G.~Kylberg and I.~Sintorn, ``Evaluation of noise robustness for local binary pattern descriptors in texture classification,'' \emph{{EURASIP} J. Image Video Process.}, vol. 2013, p.~17, 2013. [Online]. Available: \url{https://doi.org/10.1186/1687-5281-2013-17}
\BIBentrySTDinterwordspacing

\bibitem{DBLP:journals/prl/IdrissaA02}
\BIBentryALTinterwordspacing
M.~Idrissa and M.~Acheroy, ``Texture classification using gabor filters,'' \emph{Pattern Recognit. Lett.}, vol.~23, no.~9, pp. 1095--1102, 2002. [Online]. Available: \url{https://doi.org/10.1016/S0167-8655(02)00056-9}
\BIBentrySTDinterwordspacing

\bibitem{DBLP:conf/cvpr/JegouDSP10}
\BIBentryALTinterwordspacing
H.~J{\'{e}}gou, M.~Douze, C.~Schmid, and P.~P{\'{e}}rez, ``Aggregating local descriptors into a compact image representation,'' in \emph{The Twenty-Third {IEEE} Conference on Computer Vision and Pattern Recognition, {CVPR} 2010, San Francisco, CA, USA, 13-18 June 2010}.\hskip 1em plus 0.5em minus 0.4em\relax {IEEE} Computer Society, 2010, pp. 3304--3311. [Online]. Available: \url{https://doi.org/10.1109/CVPR.2010.5540039}
\BIBentrySTDinterwordspacing

\bibitem{DBLP:conf/cvpr/ZhangXD17}
\BIBentryALTinterwordspacing
H.~Zhang, J.~Xue, and K.~J. Dana, ``Deep {TEN:} texture encoding network,'' in \emph{2017 {IEEE} Conference on Computer Vision and Pattern Recognition, {CVPR} 2017, Honolulu, HI, USA, July 21-26, 2017}.\hskip 1em plus 0.5em minus 0.4em\relax {IEEE} Computer Society, 2017, pp. 2896--2905. [Online]. Available: \url{https://doi.org/10.1109/CVPR.2017.309}
\BIBentrySTDinterwordspacing

\bibitem{zhai2020deep}
W.~Zhai, Y.~Cao, Z.-J. Zha, H.~Xie, and F.~Wu, ``Deep structure-revealed network for texture recognition,'' in \emph{Proceedings of the IEEE/CVF Conference on Computer Vision and Pattern Recognition}, 2020, pp. 11\,010--11\,019.

\bibitem{chen2021deep}
Z.~Chen, F.~Li, Y.~Quan, Y.~Xu, and H.~Ji, ``Deep texture recognition via exploiting cross-layer statistical self-similarity,'' in \emph{Proceedings of the IEEE/CVF Conference on Computer Vision and Pattern Recognition}, 2021, pp. 5231--5240.

\bibitem{xu2021encoding}
Y.~Xu, F.~Li, Z.~Chen, J.~Liang, and Y.~Quan, ``Encoding spatial distribution of convolutional features for texture representation,'' \emph{Advances in Neural Information Processing Systems}, vol.~34, pp. 22\,732--22\,744, 2021.

\bibitem{zhu2023learning}
L.~Zhu, T.~Chen, J.~Yin, S.~See, and J.~Liu, ``Learning gabor texture features for fine-grained recognition,'' in \emph{Proceedings of the IEEE/CVF International Conference on Computer Vision}, 2023, pp. 1621--1631.

\bibitem{mohan2024lacunarity}
A.~Mohan and J.~Peeples, ``Lacunarity pooling layers for plant image classification using texture analysis,'' in \emph{Proceedings of the IEEE/CVF Conference on Computer Vision and Pattern Recognition}, 2024, pp. 5384--5392.

\bibitem{zhai2023exploring}
W.~Zhai, Y.~Cao, J.~Zhang, H.~Xie, D.~Tao, and Z.-J. Zha, ``On exploring multiplicity of primitives and attributes for texture recognition in the wild,'' \emph{IEEE Transactions on Pattern Analysis and Machine Intelligence}, 2023.

\bibitem{leung2001representing}
T.~Leung and J.~Malik, ``Representing and recognizing the visual appearance of materials using three-dimensional textons,'' \emph{International journal of computer vision}, vol.~43, no.~1, pp. 29--44, 2001.

\bibitem{varma2005statistical}
M.~Varma and A.~Zisserman, ``A statistical approach to texture classification from single images,'' \emph{International journal of computer vision}, vol.~62, no. 1-2, pp. 61--81, 2005.

\bibitem{varma2008statistical}
------, ``A statistical approach to material classification using image patch exemplars,'' \emph{IEEE transactions on pattern analysis and machine intelligence}, vol.~31, no.~11, pp. 2032--2047, 2008.

\bibitem{varma2003texture}
------, ``Texture classification: Are filter banks necessary?'' in \emph{2003 IEEE Computer Society Conference on Computer Vision and Pattern Recognition, 2003. Proceedings.}, vol.~2.\hskip 1em plus 0.5em minus 0.4em\relax IEEE, 2003, pp. II--691.

\bibitem{2016Facilitating}
S.~Ahmad and L.~F. Cheong, ``Facilitating and exploring planar homogeneous texture for indoor scene understanding,'' in \emph{European Conference on Computer Vision}, 2016.

\bibitem{2016Sparse}
P.~Koniusz and A.~Cherian, ``Sparse coding for third-order super-symmetric tensor descriptors with application to texture recognition,'' in \emph{Computer Vision \& Pattern Recognition}, 2016, pp. 5395--5403.

\bibitem{liu2019bow}
L.~Liu, J.~Chen, P.~Fieguth, G.~Zhao, R.~Chellappa, and M.~Pietik{\"a}inen, ``From bow to cnn: Two decades of texture representation for texture classification,'' \emph{International Journal of Computer Vision}, vol. 127, no.~1, pp. 74--109, 2019.

\bibitem{2016Equiangular}
Y.~Quan, C.~Bao, and J.~Hui, ``Equiangular kernel dictionary learning with applications to dynamic texture analysis,'' in \emph{Computer Vision \& Pattern Recognition}, 2016.

\bibitem{2015Texture}
L.~A. Gatys, A.~S. Ecker, and M.~Bethge, ``Texture synthesis using convolutional neural networks,'' \emph{MIT Press}, 2015.

\bibitem{lin2015bilinear}
T.-Y. Lin, A.~RoyChowdhury, and S.~Maji, ``Bilinear cnn models for fine-grained visual recognition,'' in \emph{Proceedings of the IEEE international conference on computer vision}, 2015, pp. 1449--1457.

\bibitem{andrearczyk2016using}
V.~Andrearczyk and P.~F. Whelan, ``Using filter banks in convolutional neural networks for texture classification,'' \emph{Pattern Recognition Letters}, vol.~84, pp. 63--69, 2016.

\bibitem{fujieda2017wavelet}
S.~Fujieda, K.~Takayama, and T.~Hachisuka, ``Wavelet convolutional neural networks for texture classification,'' \emph{arXiv preprint arXiv:1707.07394}, 2017.

\bibitem{bu2019deep}
X.~Bu, Y.~Wu, Z.~Gao, and Y.~Jia, ``Deep convolutional network with locality and sparsity constraints for texture classification,'' \emph{Pattern Recognition}, vol.~91, pp. 34--46, 2019.

\bibitem{wang2018non}
X.~Wang, R.~Girshick, A.~Gupta, and K.~He, ``Non-local neural networks,'' in \emph{Proceedings of the IEEE conference on computer vision and pattern recognition}, 2018, pp. 7794--7803.

\bibitem{DBLP:conf/cvpr/Xue0D18}
J.~Xue, H.~Zhang, and K.~J. Dana, ``Deep texture manifold for ground terrain recognition,'' in \emph{2018 {IEEE} Conference on Computer Vision and Pattern Recognition, {CVPR} 2018, Salt Lake City, UT, USA, June 18-22, 2018}.\hskip 1em plus 0.5em minus 0.4em\relax {IEEE} Computer Society, 2018, pp. 558--567.

\bibitem{lin2016visualizing}
T.-Y. Lin and S.~Maji, ``Visualizing and understanding deep texture representations,'' in \emph{Proceedings of the IEEE conference on computer vision and pattern recognition}, 2016, pp. 2791--2799.

\bibitem{peeples2020histogram}
J.~Peeples, W.~Xu, and A.~Zare, ``Histogram layers for texture analysis,'' \emph{arXiv preprint arXiv:2001.00215}, 2020.

\bibitem{cimpoi14describing}
M.~Cimpoi, S.~Maji, I.~Kokkinos, S.~Mohamed, , and A.~Vedaldi, ``Describing textures in the wild,'' in \emph{Proceedings of the {IEEE} Conf. on Computer Vision and Pattern Recognition ({CVPR})}, 2014.

\bibitem{sharan2013recognizing}
L.~Sharan, C.~Liu, R.~Rosenholtz, and E.~H. Adelson, ``Recognizing materials using perceptually inspired features,'' \emph{International journal of computer vision}, vol. 103, no.~3, pp. 348--371, 2013.

\bibitem{bell2015material}
S.~Bell, P.~Upchurch, N.~Snavely, and K.~Bala, ``Material recognition in the wild with the materials in context database,'' in \emph{Proceedings of the IEEE conference on computer vision and pattern recognition}, 2015, pp. 3479--3487.

\bibitem{Eccv16KampourisZGM}
C.~Kampouris, S.~Zafeiriou, A.~Ghosh, and S.~Malassiotis, ``Fine-grained material classification using micro-geometry and reflectance,'' in \emph{Computer Vision - {ECCV} 2016 - 14th European Conference, Amsterdam, The Netherlands, October 11-14, 2016}, B.~Leibe, J.~Matas, N.~Sebe, and M.~Welling, Eds.\hskip 1em plus 0.5em minus 0.4em\relax Springer, 2016, pp. 778--792.

\bibitem{xue2017differential}
J.~Xue, H.~Zhang, K.~Dana, and K.~Nishino, ``Differential angular imaging for material recognition,'' in \emph{Proceedings of the IEEE Conference on Computer Vision and Pattern Recognition}, 2017, pp. 764--773.

\bibitem{caputo2005class}
B.~Caputo, E.~Hayman, and P.~Mallikarjuna, ``Class-specific material categorisation,'' in \emph{Tenth IEEE International Conference on Computer Vision (ICCV'05) Volume 1}, vol.~2.\hskip 1em plus 0.5em minus 0.4em\relax IEEE, 2005, pp. 1597--1604.

\bibitem{han2022vision}
K.~Han, Y.~Wang, J.~Guo, Y.~Tang, and E.~Wu, ``Vision gnn: An image is worth graph of nodes,'' \emph{arXiv preprint arXiv:2206.00272}, 2022.

\bibitem{dosovitskiy2020image}
A.~Dosovitskiy, L.~Beyer, A.~Kolesnikov, D.~Weissenborn, X.~Zhai, T.~Unterthiner, M.~Dehghani, M.~Minderer, G.~Heigold, S.~Gelly \emph{et~al.}, ``An image is worth 16x16 words: Transformers for image recognition at scale,'' \emph{arXiv preprint arXiv:2010.11929}, 2020.

\bibitem{liu2021swin}
Z.~Liu, Y.~Lin, Y.~Cao, H.~Hu, Y.~Wei, Z.~Zhang, S.~Lin, and B.~Guo, ``Swin transformer: Hierarchical vision transformer using shifted windows,'' \emph{arXiv preprint arXiv:2103.14030}, 2021.

\end{thebibliography}

\end{document}